\documentclass[sigconf]{acmart}

\usepackage{booktabs} % For formal tables

\setcopyright{rightsretained}
\copyrightyear{2020}
\acmYear{2020}
\setcopyright{acmcopyright}
\acmConference[GECCO '20]{Genetic and Evolutionary Computation Conference}{July 8--12, 2020}{Cancun, Mexico}
\acmBooktitle{Genetic and Evolutionary Computation Conference (GECCO '20), July 8--12, 2020, Cancun, Mexico}
\acmPrice{15.00}
\acmDOI{10.1145/3377930.3389842}
\acmISBN{978-1-4503-7128-5/20/07}

\usepackage{amssymb}
\usepackage{amsmath}
\usepackage{bm}
\usepackage{mathtools}

\DeclarePairedDelimiterX{\infdivx}[2]{(}{)}{%
  #1\;\delimsize\|\;#2%
}

\renewcommand{\vec}[1]{\mathbf{#1}}

\usepackage{color}
\usepackage{xcolor}

\begin{document}

\title{Effective Reinforcement Learning through Evolutionary Surrogate-Assisted Prescription}
\author{Olivier Francon$^{1}$, Santiago Gonzalez$^{1,2}$, Babak Hodjat$^{1}$, Elliot Meyerson$^{1}$,\\
Risto Miikkulainen$^{1,2}$, Xin Qiu$^{1}$, and Hormoz Shahrzad$^{1}$}
\affiliation{
$^{1}$Cognizant Technology Solutions and $^{2}$The University of Texas at Austin\\[2ex]}
\renewcommand{\shortauthors}{Francon, Gonzalez, Hodjat, Meyerson,
  Miikkulainen, Qiu, and Shahrzad}

\begin{abstract}
There is now significant historical data available on decision making in organizations, consisting of the decision problem, what decisions were made, and how desirable the outcomes were. Using this data, it is possible to learn a surrogate model, and with that model, evolve a decision strategy that optimizes the outcomes. This paper introduces a general such approach, called Evolutionary Surrogate-Assisted Prescription, or ESP. The surrogate is, for example, a random forest or a neural network trained with gradient descent, and the strategy is a neural network that is evolved to maximize the predictions of the surrogate model. ESP is further extended in this paper to sequential decision-making tasks, which makes it possible to evaluate the framework in reinforcement learning (RL) benchmarks. Because the majority of evaluations are done on the surrogate, ESP is more sample efficient, has lower variance, and lower regret than standard RL approaches. Surprisingly, its solutions are also better because both the surrogate and the strategy network regularize the decision making behavior. ESP thus forms a promising foundation to decision optimization in real-world problems.
\end{abstract}



\maketitle

\section{Introduction}

\begin{figure}
    \centering
    \includegraphics[width=0.6\columnwidth]{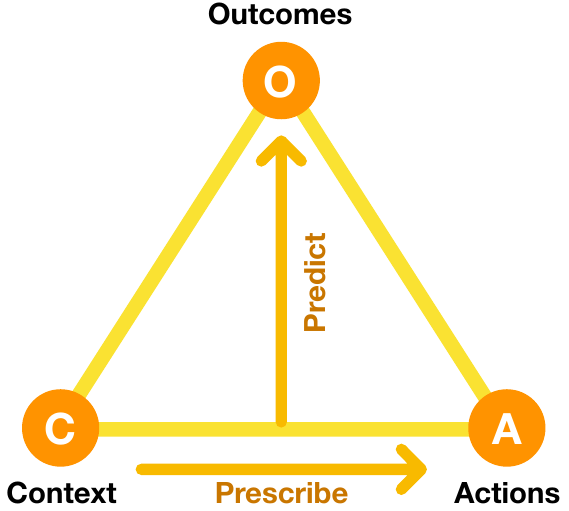}
    \caption{\emph{Elements of ESP.} A Predictor is trained with historical data on how given actions in given contexts led to specific outcomes. The Predictor can be any machine learning model trained with supervised methods, such as a random forest or a neural network. The Predictor is then used as a surrogate in order to evolve a Prescriptor, i.e.\ a neural network implementing a decision policy that results in the best possible outcomes. The majority of evaluations are done on the surrogate, making the process highly sample-efficient and robust, and leading to decision policies that are regularized and therefore generalize well.}
    \label{fg:esptriangle}
\end{figure}

Many organizations in business, government, education, and healthcare
now collect significant data about their operations. Such data is
transforming decision making in organizations: It is now possible to
use machine learning techniques to build predictive models of
behaviors of customers, consumers, students, and competitors, and, in
principle, make better decisions, i.e.\ those that
lead to more desirable outcomes. However, while prediction is
necessary, it is only part of the process.  Predictive models do not
specify what the optimal decisions actually are.  To find a good
decision strategy, different approaches are needed.

The main challenge is that optimal strategies are not known, so
standard gradient-based machine learning approaches cannot be used.
The domains are only partially observable, and decision variables and
outcomes often interact nonlinearly: For instance, allocating marketing
resources to multiple channels may have a nonlinear cumulative effect,
or nutrition and exercise may interact to leverage or undermine the
effect of medication in treating an illness \cite{dreer:behavmed17,naik:marketingsci05}. Such
interactions make it difficult to utilize linear programming and other
traditional optimization approaches from operations research.

Instead, good decision strategies need to be found using search, i.e.\
by generating strategies, evaluating them, and generating new, hopefully
better strategies based on the outcomes. In many domains such search
cannot be done in the domain itself: For instance, testing an ineffective marketing
strategy or medical treatment could be prohibitively costly; evaluating an
engineering design through simulation, or a behavioral strategy in game playing could
require a prohibitive amount of computation time. However,
given that historical data about past decisions and their outcomes
exist, it is possible to do the search using a predictive model as a
surrogate to evaluate them. Only once good decision strategies have been found using
the surrogate, they are tested in the real world.

Even with the surrogate, the problem of finding effective decision
strategies is still challenging. Nonlinear interactions may result in
deceptive search landscapes, where progress towards good solutions
cannot be made through incremental improvement: Discovering them
requires large, simultaneous changes to multiple variables.
Decision strategies often require balancing multiple objectives, such
as performance and cost, and in practice, generating a number of
different trade-offs between them is needed. Consequently, search
methods such as reinforcement learning (RL), where a solution is gradually
improved through local exploration, do not lend themselves well to
searching solution strategies either.  Further, the number of
variables can be very large, e.g.\ thousands or even millions as in
some manufacturing and logistics problems \cite{deb:ejor17}, making methods
such as Kriging and Bayesian optimization \cite{cressie:mathgeo90,snoek:nips12} ineffective.
Moreover, the solution is not a single point but a strategy, i.e.\ a
function that maps input situations to optimal decisions, exacerbating
the scale-up problem further.

Keeping in mind the above challenges, an approach is developed in this
paper for \emph{Evolutionary Surrogate-Assisted Prescription} (ESP; Figure~\ref{fg:esptriangle}), i.e.\
for discovering effective solution strategies using evolutionary
optimization. With a population-based search method, it is possible to
navigate deceptive, high-dimensional landscapes, and discover
trade-offs across multiple objectives \cite{miikkulainen:action18}. The
strategy is expressed as a neural network, making it possible to use
state-of-the-art neuroevolution techniques to optimize it. Evaluations
of the neural network candidates are done using a predictive model,
trained with historical data on past decisions and their outcomes.

Elements of the ESP approach were already found effective in
challenging real-world applications. In an autosegmentation version of
Ascend by Evolv, a commercial product for optimizing designs of web
pages \cite{miikkulainen:aimag20}, a neural network was used to map user descriptions
to most effective web-page designs. In CyberAg, effective growth
recipes for basil were found through search with a surrogate model
trained with outcomes of past recipes \cite{johnson:plos19}. In both cases,
evolutionary search found designs that were more effective than human
designs, even surprising and unlikely to be found by humans. In ESP, these elements of strategy search and surrogate modeling are combined into a general approach for decision strategy optimization. 
ESP is implemented as part of LEAF, i.e.\ Cognizant's Evolutionary AI
platform, and is currently being applied to a number of business
decision optimization problems.

The goal of this paper is to introduce the ESP approach in general, and to extend it further into decision
strategies that consist of sequences of decisions. This extension
makes it possible to evaluate ESP against other methods in RL domains.
Conversely, ESP is used to formalize RL as
surrogate-assisted, population based search. This approach is
particularly compelling in domains where real-world evaluations are
costly. ESP improves upon traditional RL in several ways: It
converges faster given the same number of episodes, indicating better sample-efficiency;
it has lower variance for best policy performance, indicating better reliability of delivered solutions; and it has lower regret, indicating lower costs and better safety during training.
Surprisingly, optimizing against the surrogate
also has a regularization effect: the solutions are sometimes more
general and thus perform better than solutions discovered in the
domain itself. Further, ESP brings the advantages of population-based
search outlined above to RL, i.e.\ enhanced
exploration, multiobjectivity, and scale-up to high-dimensional
search spaces.

The ESP approach is evaluated in this paper in various
RL benchmarks. First its behavior is visualized in a synthetic domain, illustrating how the Predictor and Prescriptor learn together to discover optimal decisions. Direct evolution (DE) is compared to evolution with the surrogate, to demonstrate how the approach minimizes the need for evaluations in the real world. ESP is then compared with standard RL approaches, demonstrating better sample efficiency, reliability, and lower cost. The experiments also demonstrate regularization through the surrogate, as well as ability to utilize different kinds of Predictor models (e.g.\ random forests and neural networks) to best fit the data. ESP
thus forms a promising evolutionary-optimization-based approach to
sequential decision-making tasks.

\section{Related work}

Traditional model-based RL aims to build a transition model, embodying the system's dynamics, that estimates the system's next state in time, given current state and actions. The transition model, which is learned as part of the RL process, allows for effective action selection to take place \cite{Ha2018}. These models allow agents to leverage predictions of the future state of their environment \cite{werbos1987learning}. However, model-based RL usually requires a prohibitive amount of data for building a reliable transition model while also training an agent. Even simple systems can require tens- to hundreds-of-thousands of samples \cite{schmidhuber1991learning}. While techniques such as PILCO \cite{deisenroth2011pilco} have been developed to specifically address sample efficiency in model-based RL, they can be computationally intractable for all but the lowest dimensional domains \cite{wahlstrom2015pixels}.

As RL has been applied to increasingly complex tasks with real-world
costs, sample efficiency has become a crucial issue. Model-free RL has
emerged as an important alternative in such tasks because they sample
the domain without a transition model. Their performance and efficiency thus depend
on their sampling and reward estimation methods. As a representative model-free
off-policy method, Deep Q-Networks (DQN) \cite{mnih2015dqn} solves the
sample efficiency issue by modeling future rewards using action values, also
known as Q values. The Q-network is learned based on a replay buffer that
collects training data from real-world interactions. Advanced techniques such as double Q-learning \cite{Hasselt2010} and dueling
network architectures \cite{Wang2016} makes DQN competitive
in challenging problems. 

In terms of on-policy model-free
techniques, policy gradient approaches (sometimes
referred to as deep RL) leverage developments in deep neural networks to provide a general RL
solution. Asynchronous Advantage Actor-Critic (A3C)
\cite{mnih2016asynchronous} in particular builds policy critics through an advantage
function, which considers both action and state values. Proximal Policy
Optimization (PPO) \cite{Schulman2017} further makes actor-critic methods
more robust with a clipped surrogate objective. Unfortunately,
policy gradient techniques are typically quite sensitive to hyperparameter
settings, and require overwhelming numbers of samples. One reason 
is the need to train a policy neural network. While
expensive, policy gradient methods have had success in simulated 3D
locomotion \cite{schulman2015high, heess2017emergence}, Atari video games
\cite{mnih2015dqn}, and, famously, Go \cite{silver2016mastering}. 

ESP will be compared to both DQN and PPO methods in this paper. Compared to ESP, these existing RL approaches have several limitations: First, their performance during real-world interactions cannot be guaranteed, leading to safety concerns \cite{Ray2019}. In ESP, only elite agents selected via the surrogate model are evaluated on the real world, significantly improving safety. Second, the quality of the best-recognized policy is unreliable because it has not been sufficiently evaluated during learning. ESP solves this issue by evaluating all elite policies in the real world for multiple episodes. Third, existing RL methods rely heavily on deep neural networks. In contrast, ESP treats the Predictor as a black box, allowing high flexibility in model choices, including simpler models such as random forests that are sufficient in many cases.

%The above RL approaches are all known as ``model-free'' RL due to the fact that they sample without a model of the system. Model-free approaches performance and data-efficiency thus depend on the sampling methodology and reward estimation that are used. Model-based RL aims to tackle this by including a model of the system's dynamics that estimates the system's next state in time, given a state and actions. The dynamics model, which is learned as part of the RL process, allows for more effective action selection to take place \cite{Ha2018}.
%\TODO[Since ESP is actually trying to build the reward function, not the world model, maybe we should only briefly talk about the model-based RL approaches at the beginning, and emphasize they are generally less sample-efficient than model-free approaches.]

%- Like Neural Fitted Q iteration \cite{riedmiller2005neural}

Evolutionary approaches have been used in RL in several ways.
For instance, in evolved policy gradients \cite{houthooft2018evolved}, the
loss function against which an agent's policy is optimized is evolved. The
policy loss is not represented symbolically, but rather as a neural network
that convolves over a
temporal sequence of context vectors; the parameters to this neural network
are optimized using evolutionary strategies. In reward function search \cite{niekum2010genetic}, the task is framed as a genetic programming problem,
leveraging PushGP \cite{push}.

More significantly, entire policies can be discovered directly with many evolutionary techniques including Genetic Algorithms, Natural Evolutionary Strategies, and Neuroevolution
\cite{salimans2017,such2017deepga,stanley:naturemi19,ackley1991interactions, gomez2006efficient, whiteson2012evolutionary}.
Such direct policy evolution can only leverage samples from
the real-world, which is costly and risky since many evaluations have to take place in low-performing areas of
the policy space. This issue in evolutionary optimization led
to the development of surrogate-assisted evolution techniques \cite{grefenstette1985genetic,jin2011, jin:ieeetec19}. Surrogate methods 
have been applied to a wide range of
domains, ranging from turbomachinery blade optimization
\cite{pierret1999turbomachinery} to protein design
\cite{schneider1994artificial}.  ESP aims to build upon this idea, by relegating the majority of
policy evaluations to a flexible surrogate model, allowing for a wider variety
of contexts to be used to evolve better policies.

\section{The ESP Approach}
\label{sc:approach}
The goal of the ESP approach is to find a decision policy that optimizes a set of outcomes (Figure~\ref{fg:esptriangle}). Given a set of possible contexts (or states) $\mathbb{C}$ and possible actions $\mathbb{A}$, a decision policy $D$ returns a set of actions $A$ to be performed in each context $C$:
\begin{equation}
    D(C) = A\;, 
\end{equation}
where $C \in \mathbb{C}$ and $A \in \mathbb{A}$. For each such $(C,A)$ pair there is a set of outcomes $O(C,A)$, i.e.\ the consequences of carrying out decision $A$ in context $C$. For instance, the context might be a description of a patient, actions might be medications, and outcomes might be health, side effects, and costs. In the following, higher values of $O$ are assumed to be better for simplicity.

In ESP, two models are employed: a Predictor $P_d$, and a Prescriptor $P_s$. The Predictor takes, as its input, context information, as well as actions performed in that context. The output of the Predictor is the resulting outcomes when the given actions are applied in the given context. The Predictor is therefore defined as
\begin{equation}
P_d (C, A) = O',
\end{equation}
such that $\sum_j L(O_j,O_j')$ across all dimensions $j$ of $O$ is minimized. The function $L$ can be any of the usual loss functions used in machine learning, such as cross-entropy or mean-squared-error, and the model $P_d$ itself can be any supervised machine learning model such as a neural network or a random forest.

The Prescriptor takes a given context as input, and outputs a set of actions:
\begin{equation}
P_s (C) = A\;,
\end{equation}
such that $\sum_{i,j} O_j'(C_i,A_i)$ over all possible contexts $i$ is maximized. It thus approximates the optimal decision policy for the problem. Note that the optimal actions $A$ are not known, and must therefore be found through search.

\begin{figure}
    \centering
    \includegraphics[width=\columnwidth]{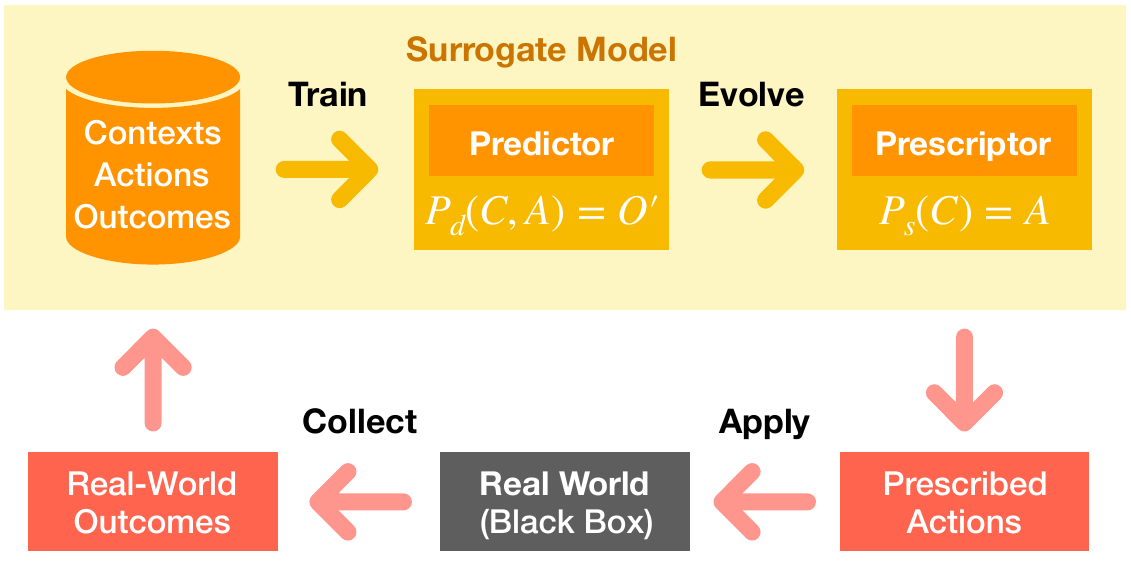}
    \caption{\emph{The ESP Outer Loop.} The Predictor can be trained gradually at the same time as the Prescriptor is evolved, using the Prescriptor to drive exploration. That is, the user can decide to apply the Prescriptor's outputs to the real world, observe the outcomes, and aggregate them into the Predictor's training set.}
    \label{fg:outerloop}
\end{figure}

The ESP algorithm then operates as an outer loop that constructs the Predictor and Prescriptor models (Figure~\ref{fg:outerloop}):
\begin{enumerate}
\setlength{\itemsep}{0ex}
\item\label{st:predictor} Train a Predictor based on historical training data;
\item Evolve Prescriptors with the Predictor as the surrogate;
\item Apply the best Prescriptor in the real world;
\item\label{st:newdata} Collect the new data and add to the training set;
\item Repeat until convergence.
\end{enumerate}
As usual in evolutionary search, the process terminates when a satisfactory level of outcomes has been reached, or no more progress can be made. Note that in Step~\ref{st:predictor}, if no historical decision data exists initially, a random Predictor can be used.
Also note that not all data needs to be accumulated for training each iteration. In domains where the underlying relationships between variables might change over time, it might be advisable to selectively ignore samples from the older data as more data is added to the training set in Step~\ref{st:newdata}. It is thus possible to bias the training set towards more recent experiences.

\begin{figure*}[t]
    \centering
    \includegraphics[width=7in]{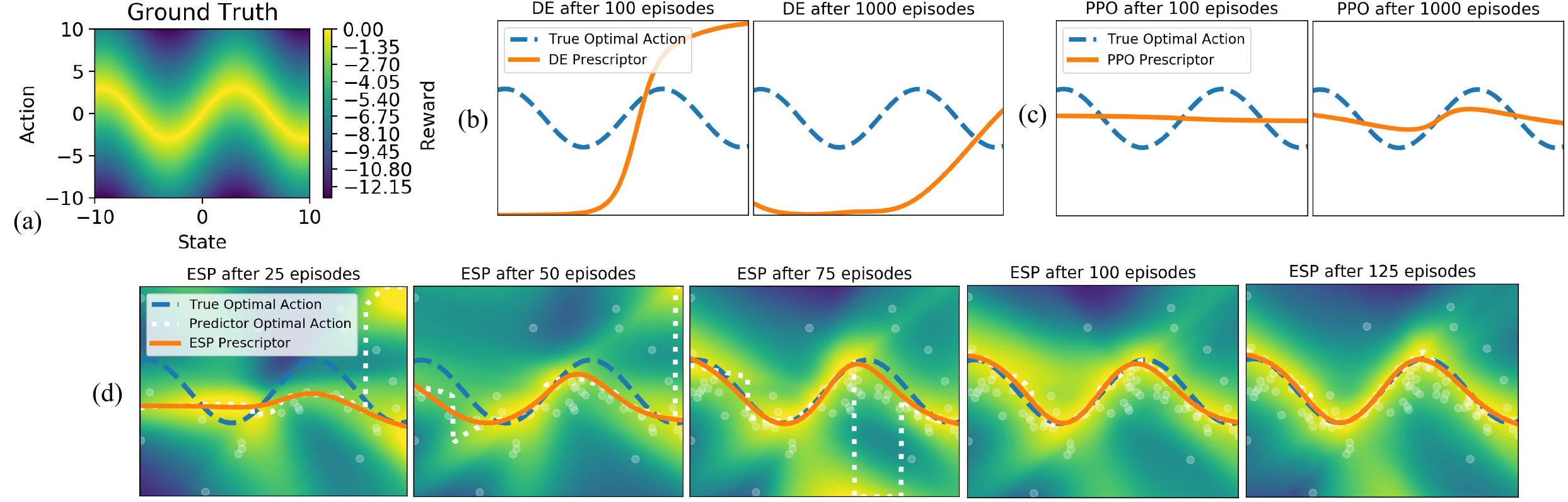}
    \caption{\emph{Visualizing ESP Behavior.} 
    $(a)$ True reward for every state-action pair;
    $(b-c)$ After 1000 episodes, the top Prescriptors for direct evolution (DE) and PPO are still far from optimal;
    $(d)$ ESP Prescriptor (orange) and Predictor (background) for several iterations.
    The translucent circles indicate the state-action pairs sampled so far, i.e., the samples on which the Predictor is trained.
    By 125 episodes, ESP has converged around the optimal Prescriptor, and the ESP Predictor has converged in the neighborhood of this optimum, showing how ESP can leverage Predictors over time to find good actions quickly.
    Note that the Prescriptor does not exactly match the actions the Predictor would suggest as optimal: the Prescriptor regularizes the Predictor's overfitting by implicitly ensembling the Predictors evolved against over time.
    For a full video of the algorithms, see \url{http://evolution.ml/demos/espvisualization}.
    }
    \label{fig:vis_frames}
\end{figure*}

Building the Predictor model is straightforward given a $(C,A,O)$ dataset. The choice of algorithm depends on the domain, i.e.\ how much data there is, whether it is continuous or discrete, structured or unstructured. Random forests and neural networks will be demonstrated in this role in this paper. The Prescriptor model, in contrast, is built using neuroevolution in ESP: Neural networks because they can express complex nonlinear mappings naturally, and evolution because it is an efficient way to discover such mappings \cite{stanley:naturemi19}, and naturally suited to optimize multiple objectives \cite{coello:ijkais99,emmerich:natcomp18}. Because it is evolved with the Predictor, the Prescriptor is not restricted by a finite training dataset, or limited opportunities to evaluate in the real world. Instead, the Predictor serves as a fitness function, and it can be queried frequently and efficiently. %As a standard neuroevolution method, NEAT \cite{stanley:ec02} will be used to evolve the weights of the Prescriptor neural network in this paper.
In a multiobjective setting, ESP produces multiple Prescriptors, selected from the Pareto front of the multiobjective neuroevolution run.

Applying the ESP framework to RL problems involves extending the contexts and actions to sequences. The Prescriptor can be seen as an RL agent, taking the current context as input, and deciding what actions to perform in each time step. The output of the Predictor, $O'$, can be seen as the reward vector for that step, i.e.\ as Q values \cite{watkins1989qlearning} (with a given discount factor, such as $0.9$, as in the experiments below). Evolution thus aims to maximize the predicted reward, or minimize the regret, throughout the sequence. 

The outer loop of ESP changes slightly because in RL there is no dataset to train the Predictor; instead, the data needs to be generated by applying the current Prescriptors to the domain. An elite set of several good Prescriptors are used in this role to create a more diverse training set. The initial training set is created randomly. The loop now is:
\begin{enumerate}
\setlength{\itemsep}{0ex}
\item Apply the elite Prescriptors in the actual domain;
\item\label{st:qvalues} Collect Q values for each time step for each Prescriptor;
\item Train a Predictor based on data collected in Step~\ref{st:qvalues};
\item Evolve Prescriptors with the Predictor as the surrogate;
\item Repeat until convergence.
\end{enumerate}
The evolution of Prescriptors continues in each iteration of this loop from where it left off in previous iteration. In addition, the system keeps track of the best Prescriptor so far, as evaluated in the actual domain, and makes sure it stays in the parent population during evolution. This process discovers good Prescriptor agents efficiently, as will be described in the experiments that follow.

\section{Experiments}

ESP was evaluated in three domains: Function approximation, where its behavior could be visualized concretely; Cart-pole control \cite{barto1983cartpole}
where its performance could be compared to standard RL methods in a standard RL benchmark task; and Flappy Bird, where the regularization effect of the surrogate could be demonstrated most clearly.  

The neuroevolution algorithm for discovering Prescriptors evolves weights of neural networks with fixed topologies.
Unless otherwise specified, all experiments use the following default setup for evolution:
candidates have a single hidden layer with bias and tanh activation;
the initial population uses orthogonal initialization of layer weights with a mean of 0 and a standard deviation of 1 \cite{saxe2013orthogonalinit};
the population size is 100;
the top 10\% of the population is carried over as elites;
parents are selected by tournament selection of the top 20\% of candidates;
recombination is performed by uniform crossover at the weight-level;
there is a 10\% probability of multiplying each weight by a mutation factor, where mutation factors are drawn from $\mathcal{N}(1, 0.1)$.

\subsection{Visualizing ESP Behavior}
\label{subsec:exp_visualizing_esp}

This section demonstrates the behavior of ESP in a synthetic function approximation domain where its behavior can be visualized. The domain also allows comparing ESP to direct evolution in the domain, as well as to PPO, and visualizing their differences.

\paragraph{Problem Description}
The domain has a one-dimensional context $C$ and a one-dimensional action $A$, with outcome $O$ given by the function $O = - \lvert A - 3 \sin \frac{C}{2} \rvert$.
The optimal action for each context lies on a periodic curve, which captures complexity that can arise from periodic variables such as time of day or time of year.
The outcome of each action decreases linearly as the action moves away from the optimal action.
Episodes in this domain consist of single action in $[-10, 10]$, which is taken in a context drawn uniformly over $[-10, 10]$.
The full domain is shown in Figure~\ref{fig:vis_frames}(a).

\paragraph{Algorithm Setup}
ESP begins by taking ten random actions.
Thereafter, every iteration, ESP trains a Predictor neural network with two hidden layers of size 64 and tanh activation for 2000 epochs using the Adam optimizer \cite{adam} with default parameters to minimize MSE, and evolves Prescriptor neural networks for 20 generations against this Predictor. 
Then, the top Prescriptor is run in the real domain for a single episode.
Prescriptors have a single hidden layer of size 32 with tanh activation; default parameters are used for evolution. 

Direct Evolution (DE) was run as a baseline comparison for ESP.
It consists of the exact same evolution process, except that it is run directly against the real function instead of the Predictor. That is, in each generation, all 100 candidates are evaluated on one episode from the real function.

PPO was run as an RL comparison, since it is a state-of-the-art RL approach for continuous action spaces \cite{Schulman2017}.
During each iteration it was run for ten episodes, since this setting was found to perform best during hyperparameter search.
PPO defaults\footnote{\url{https://stable-baselines.readthedocs.io/en/master/modules/ppo2.html}} were used for the remaining hyperparameters.

Ten independent runs were performed for each algorithm.
The returned policy at any time for DE and ESP is the candidate with the highest fitness in the population; for PPO it is the learned policy run without stochastic exploration.

\paragraph{Qualitative Results}
Snapshots of the convergence behavior for each method are shown in Figures~\ref{fig:vis_frames}(b-d).
After 1000 episodes, neither DE nor PPO converged near the optimal solution.
On the other hand, ESP discovered the periodic nature of the problem within 50 episodes, and converged almost exactly to the optimal within 125 episodes.

The Predictor's predicted reward for each state-action pair is shown using the background colors in each snapshot of ESP.
The rapid convergence of the Predictor highlights the sample efficiency of ESP, due to aggressive use of historical data (shown as translucent circles).
Note, however, that the Predictor does not converge to the ground truth over the entire domain; it does so just in the neighborhood of the optimal Prescriptor.
Thus, ESP avoids excessive costly exploration of low-quality actions once the structure of optimal actions has become clear.

Note also that the Prescriptor does not follow the optimal action suggested by the Predictor at every iteration exactly.
Since it maps states directly to actions, the Prescriptor provides a smoothing regularization in action space that can overcome Predictor overfitting.
Also, since the top ESP Prescriptors must survive across many different Predictors over time, ESP benefits from an implicit temporal ensembling, which further improves regularization.

\begin{figure}
    \centering
    \begin{minipage}{0.49\columnwidth}
    \centering
    \includegraphics[width=\columnwidth]{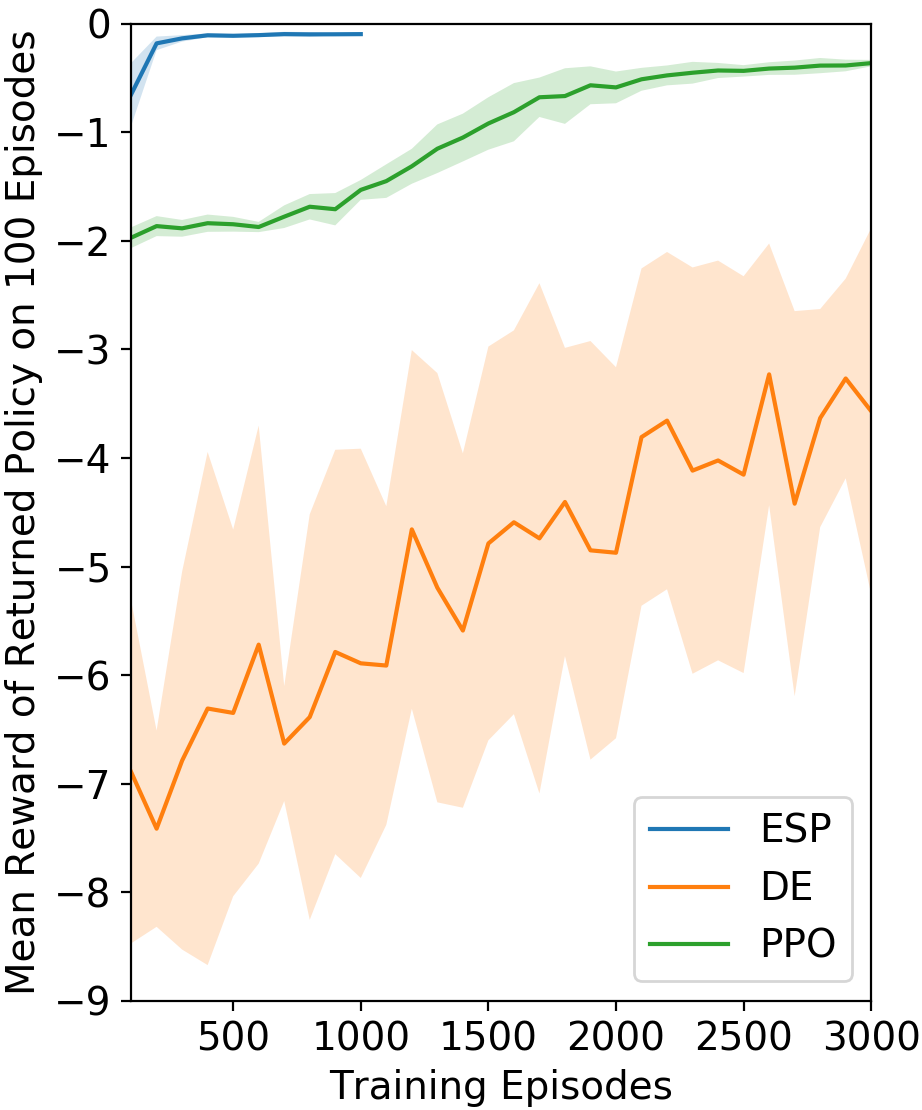}\\
    $(a)$ True Performance
    \end{minipage}
    \hfill
    \begin{minipage}{0.49\columnwidth}
    \centering
    \includegraphics[width=\columnwidth]{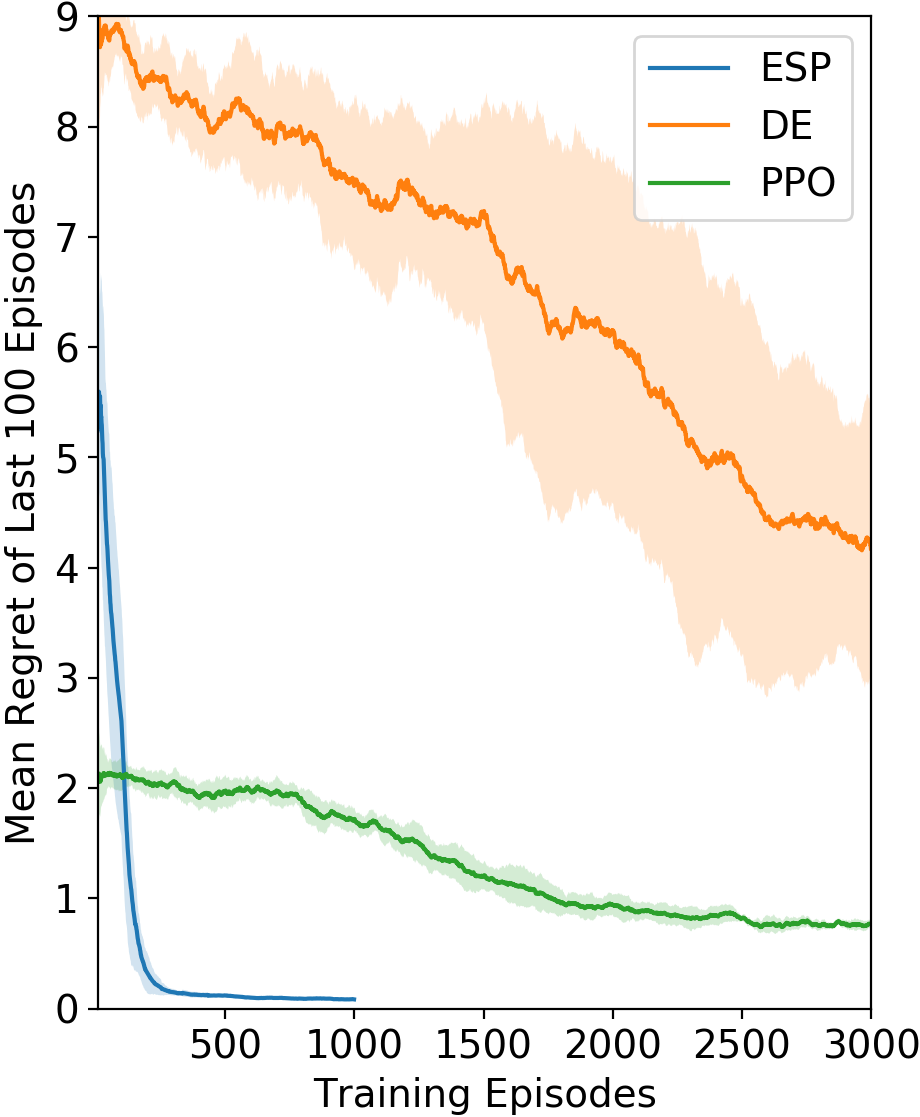}\\
    $(b)$ Regret
    \end{minipage}
    \caption{\emph{Performance in the Function Approximation Domain.}
    The horizontal axis indicates the total number of real-world episodes used by training and the vertical axis the performance at that point. Ten independent runs were performed for each method. Solid lines represent the mean over 10 runs, and colored areas show the corresponding standard deviation. 
    $(a)$ The true performance of the returned best agent converges substantially faster with ESP.
    $(b)$ ESP also operates with much lower regret than DE or PPO,
    converging to very low regret behavior orders-of-magnitude faster than the other approaches (the plot shows the moving average regret over the last 100 training episodes).
    The standard deviation of ESP is small in both metrics, attesting to the reliability of the method.
    }
    \label{fig:vis_performance}
\end{figure}

\paragraph{Quantitative Results}
The numerical performance results in Figure~\ref{fig:vis_performance} confirm the substantial advantage of ESP.
ESP converges rapidly to a high-performing returned solution (Figure~\ref{fig:vis_performance}(a)), while taking actions with significantly lower regret (defined as the reward difference between optimal solution and current policy) than the other methods during learning (Figure~\ref{fig:vis_performance}(b)).
On both metrics, ESP converges orders-of-magnitude faster than the other approaches.
In particular, after a few hundred episodes, ESP reaches a solution that is significantly better than any found by DE or PPO, even after 3,000 episodes.
This result shows that, beyond being more sample efficient, by systematically exploiting historical data, ESP is able to find solutions that direct evolution or policy gradient search cannot.

The next sections show how these advantages of ESP can be harnessed in standard RL benchmarks, focusing on the advantage of surrogate modeling over direct evolution in the domain, comparing to DQN and PPO, and demonstrating the regularization effect of the surrogate.

\iffalse
Contextual regret after $T$ real-world steps
\begin{equation}
    \label{eq:regret}
    \text{regret}(T) = \sum_{t=1}^{T} f(a_t^*, \vec{c}_t) - f(a_t, \vec{c}_t),
\end{equation}
where $a_t^*$ is an optimal action for context $\vec{c}_t$, i.e.,
\begin{equation}
    \label{eq:optimal_action}
    a_t^* = \text{argmax}_a f(a,\vec{c}_t).
\end{equation}
\fi

\begin{figure*}[t]
\begin{minipage}{\columnwidth}
	\centering
	\includegraphics[width=.98\linewidth]{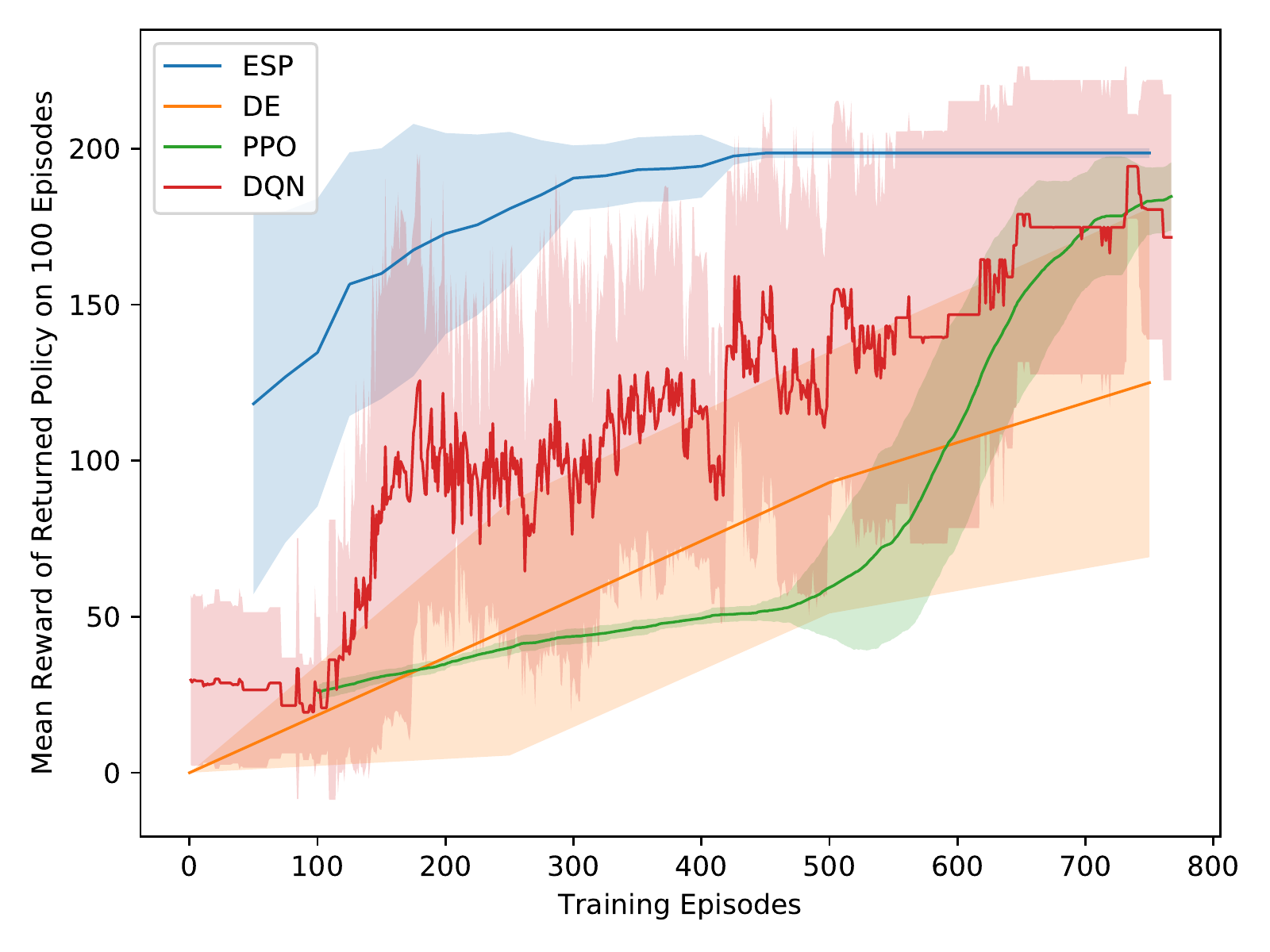}
	$(a)$ True Performance
\end{minipage}
\begin{minipage}{\columnwidth}
    \centering
	\includegraphics[width=.98\linewidth]{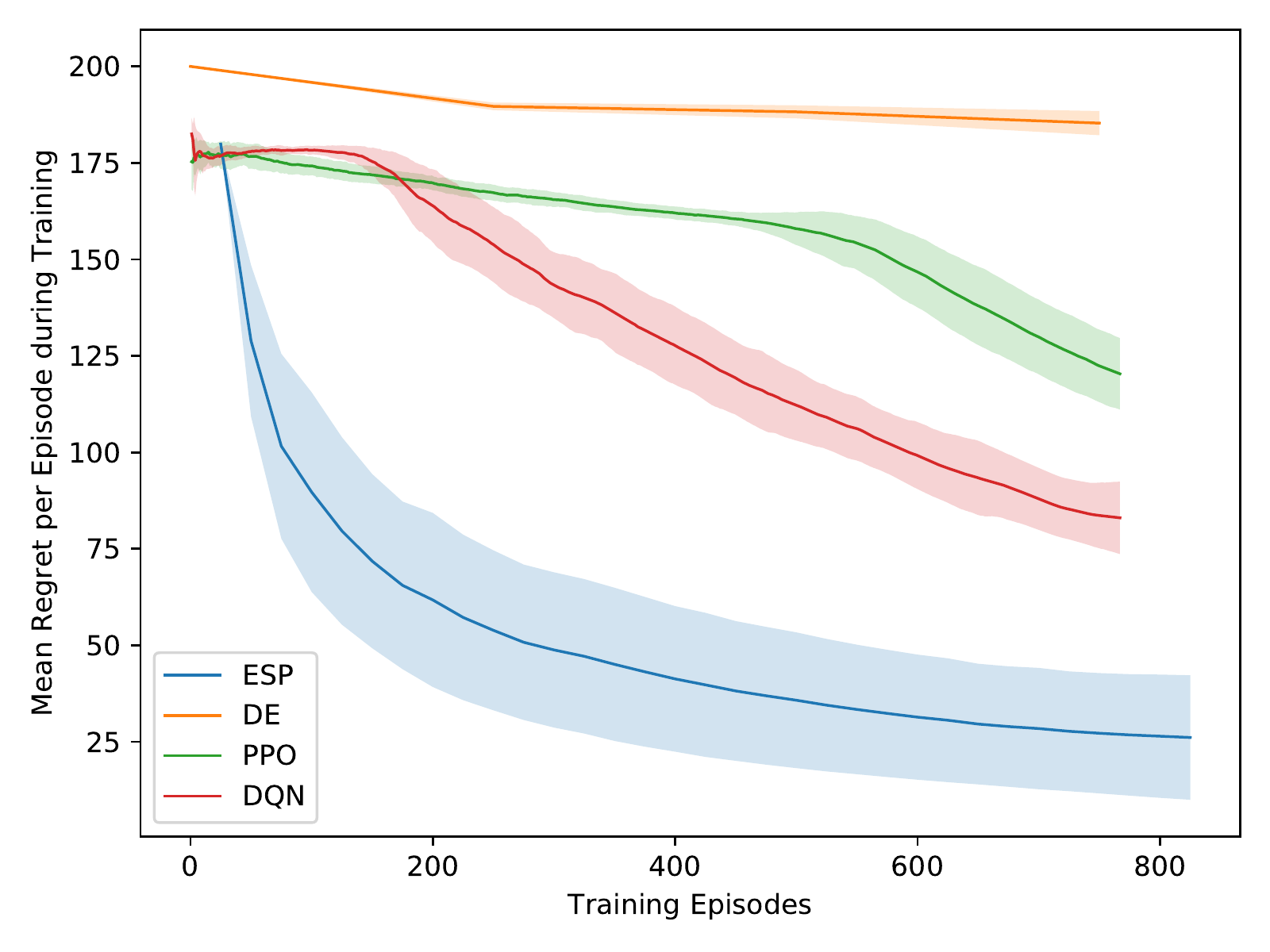}
	$(b)$ Regret
\end{minipage}
	\caption{\emph{Performance in the Cart-Pole Domain.} \label{fig:Performance_all_CartPole-v0}
    The same experimental and plotting conventions were used as in Figure~\ref{fig:vis_performance}.
    $(a)$ True performance of the best policy returned by each method during the learning process. True performance is based on the average reward of 100 real-world episodes (this evaluation is not part of the training). ESP converges significantly faster and has a much lower variance, demonstrating better sample efficiency and reliability than the other methods.
	$(b)$ Average regret over all past training episodes. ESP has significantly lower regret, suggesting that it has lower cost and is safer in real-world applications.
	}
\end{figure*}

\subsection{Comparing with Standard RL}
\label{sc:surrogate_vs_evolution}

The goal of the Cart-pole experiments was to demonstrate ESP's performance compared to direct evolution and standard RL methods. 

\paragraph{Problem Description}
The Cart-pole control domain is one of the standard RL benchmarks. In the popular CartPole-v0 implementation on the OpenAI Gym platform \cite{brockman2016gym} used in the experiments, there is a single pole on a cart that moves left and right depending on the force applied to it. A reward is given for each time step that the pole stays near vertical and the cart stays near the center of the track; otherwise the episode ends.

%The same evolution parameters were used for all the experiments. The goal was to use a generic methodology able to solve the domains without fine-tuning.\\

\paragraph{Algorithm Setup}
DE was run with a population size of 50 candidates. A candidate is a neural network with four inputs (observations), one hidden layer of 32 units with tanh activation, and two outputs (actions) with argmax activation functions.
%All layers use a bias vector. 
%The seed population consists of candidates initialized orthogonally \cite{saxe2013orthogonalinit}.
The fitness of each candidate is the average reward over five episodes in the game, where the maximum episode length is 200 time steps. %Parents are selected with tournament selection on the top 10 candidates, and five elites are carried over to the next generation unchanged. Crossover is done uniformly at the weight and bias level. Mutation probability is 10\% for each weight and bias, and the mutation extent is drawn from a normal distribution of mean 0 and standard deviation 0.1.

ESP runs similarly to DE, except that the fitness of each candidate is evaluated against the Predictor instead of the game. The Predictor is a neural network with six inputs (four observations and two actions), two hidden layers with 64 units each and tanh activation, and one output (the predicted discounted future reward) with tanh activation. It is trained for 1,000 epochs with the Adam optimizer \cite{adam} with MSE loss and batch size of 256.

The first Predictor is trained on samples collected from five random agents playing five episodes each. Random agents choose a uniform random action at each time step. A sample corresponds to a time step in the game and comprises four observations, two actions, and the discounted future reward. Reward is $+1$ on each time step, except for the last one where it is adjusted to $+2,000$ in case of success, $-2,000$ in case of failure (i.e.\ 10$\times$ max time steps). The discount factor is set to 0.9. The reward value is then scaled to lie between -1 and 1.

In order to be evaluated against the Predictor, a Prescriptor candidate has to prescribe an action for each observation vector from the collected samples. The action is then concatenated with the observation vector and passed to the Predictor to get the predicted future reward. The fitness of the candidate is the average of the predicted future rewards.

Every five generations, data is collected from the game from the five elites, for five episodes each. The new data is aggregated into the training set and a new Predictor is trained. The generation's candidates are then evaluated on the new Predictor with the new training data. The top elite candidate is also evaluated for 100 episodes on the game for reporting purposes only. Evolution is stopped after 160 generations, which corresponds to 800 episodes played from the game, or once an elite receives an average reward of 200 on five episodes. 

In addition to DE and ESP, two state-of-the-art RL methods were implemented for comparison: double DQN with dueling network architectures \cite{mnih2015dqn,Wang2016} and actor-critic style PPO \cite{Schulman2017}. The implementation and parametric setup of DQN and PPO were based on OpenAI Baselines \cite{baselines}. 
For PPO, the policy's update frequency was set to 20, which was found to be optimal during hyperparameter search.
All other parametric setups of DQN and PPO utilized default setups as recommended in OpenAI Baselines.

\paragraph{Results}
Figure~\ref{fig:Performance_all_CartPole-v0}(a) shows how the true performance of the best policy returned by ESP, DE, PPO, and DQN changes during the learning process in CartPole-v0. For ESP and DE, the elite candidate that has the best real-world fitness is selected as the best policy so far. For DQN and PPO, whenever the moving average reward of the past 100 episodes of training is increased, the best policy will be updated using the most recent policy. One hundred additional real-world episodes were used to evaluate the best policies (these evaluations are not part of the training). 

ESP converges significantly faster than the other methods, implying better sample-efficiency during learning. Moreover, the variance of the true performance becomes significantly smaller for ESP after an early stage, while all other algorithms have high variances even during later stages of learning. This observation demonstrates that the solutions delivered by ESP are highly reliable.

Figure~\ref{fig:Performance_all_CartPole-v0}(b) shows the average regret for training processes of all algorithms in CartPole-v0. ESP has significantly lower regret during the entire learning process, indicating not only lower costs but also better safety in real-world interactions.

\subsection{Regularization Through Surrogate Modeling}
\label{sc:regularization}

\begin{figure}[t]
	\centering
	%\hspace{10pt}
	\includegraphics[width=.7\linewidth]{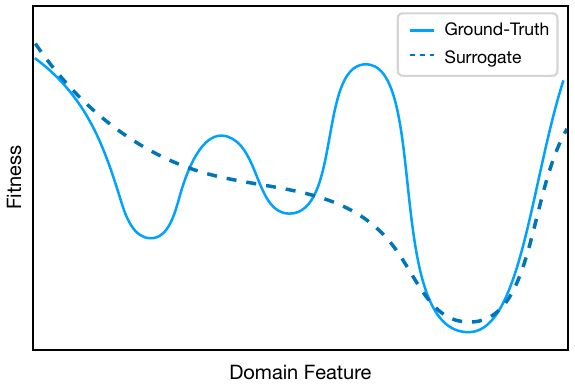}
	\caption{\emph{Surrogate Approximation of the Fitness Landscape.} \label{fig:surrogate_ref}
		The fitness in the actual domain may be deceptive and nonlinear, for instance single actions can have large consequences. The surrogate learns to approximate such a landscape, thereby creating a surrogate landscape that is easier to search and optimize.
	}
\end{figure}

Whenever a surrogate is used to approximate a fitness function, there is a risk that the surrogate introduces false optima and misleads the search \cite{jin2000} (for an inspiring collection of similar empirical phenomena, reference Lehman et al.~\cite{lehman:arxiv18}.) ESP mitigates that risk by alternating between actual domain evaluations and the surrogate. 
However, the opposite effect is also possible: Figure~\ref{fig:surrogate_ref} shows how the surrogate may form a more regularized version of the fitness than the real world, and thereby make it easier to learn policies that generalize well \cite{jin2011,ong2003}.

\paragraph{Problem Description}
Flappy Bird is a side-scroller game where the player controls a bird, attempting to fly it between columns of pipes without hitting them by performing flapping actions at carefully chosen times.
This experiment is based on a PyGame~\cite{tasfi2016PLE} implementation of this game, running at a speed of 30 frames per second. The goal of the game is to finish ten episodes of two minutes, or 3,600 frames each, through random courses of pipes.
A reward is given for each frame where the bird does not collide with the boundaries or the pipes; otherwise the episode ends. The score of each candidate is the average reward over the ten episodes.

\paragraph{Algorithm Setup}
Both DE and ESP were setup in a similar way as in the preceding sections.
DE had a population of 100 candidates, each a neural network with eight inputs (observations), one hidden layer of 128 nodes with tanh activation, and two outputs (actions) with argmax activation. %All layers had a bias vector as well.
The ESP Predictor was a random forest with 100 estimators, approximating reward values for state-action pairs frame by frame. The state-action pairs were collected with the ten best candidates of each generation running ten episodes on the actual game, for the total of a hundred episodes per generation.

\begin{figure}[t]
	\centering
	\includegraphics[width=.98\linewidth]{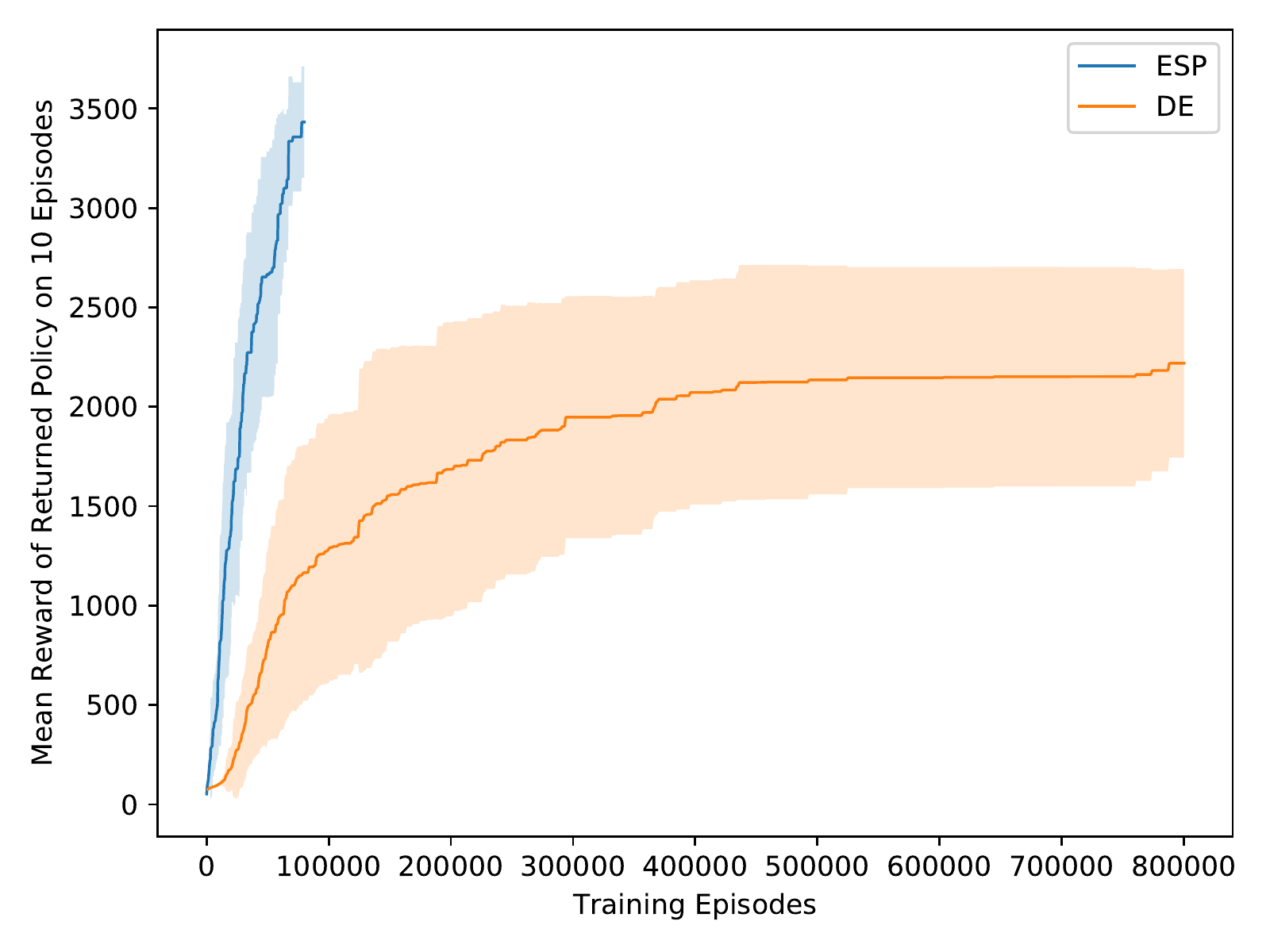}
	\caption{\emph{Performance in the Flappy Bird Domain.} \label{fig:True_performance_all_Flappy-Bird}
    The same experimental and plotting conventions were used as in 
    Figure~\ref{fig:Performance_all_CartPole-v0}(a), except true performance of the best policy returned is based on the average reward of 10 real-world episodes during the training. ESP discovers a policy that solves the task very quickly, whereas DE cannot discover it even though it is run an order of magnitude longer. This result is likely due to the regularization effect that the surrogate provides as shown in Figure~\ref{fig:surrogate_ref}.
	}
\end{figure}

\paragraph{Results} Figure~\ref{fig:True_performance_all_Flappy-Bird} shows how the true performance of the best policy returned by ESP and DE improved during the learning process. The elite candidate that has the best real-world fitness was selected as the best policy so far. In about 80,000 episodes, ESP discovered a policy that solved the task, i.e.\ was able to guide the bird through the entire course of pipes without hitting any of them. It is interesting that DE converged to a suboptimal policy even though it was run an order of magnitude longer. This result is likely due to the regularization effect illustrated in Figure~\ref{fig:surrogate_ref}. Direct evolution overfits to the nonlinear effects in the game, whereas the surrogate helps smooth the search landscape, thereby leading evolution to policies that perform better.

%\subsection{Multiobjective RL (Olivier, Santiago, Xin)}
%\begin{note}
%We'll include this if we have time to implement multiobj surrogate \\
%- in lunar lander and/or in mountain car\\
%- are there any comparisons to RL? Or just verbal comparison?\\
%- otherwise argue that this is first true multiobj RL\\
%\end{note}
 
\section{Discussion and Future Work}

The results in this paper show that the ESP approach performs well in sequential decision making tasks like those commonly used as benchmarks for RL. Compared to direct evolution and state-of-the-art RL methods, it is highly sample efficient, which is especially important in domains where exploring with the real world is costly. Its solutions are also reliable and safe, and the complexity of its models can be adjusted depending on the complexity of the data and task. These advantages apply to ESP in general, including decision strategies that are not sequential, which suggests that it is a good candidate for improving decision making in real-world applications, including those in business, government, education, and healthcare. 

When ESP is applied to such practical problems, the process outlined in Section~\ref{sc:approach} can be extended further in several ways. First, ESP can be most naturally deployed to augment human decision making. The Presciptor's output is thus taken as advice, and the human decision maker can modify the actions before applying them. These actions and their eventual outcomes are still captured and processed in Step~\ref{st:newdata} of the ESP process, and thus become part of the learning (Figure~\ref{fg:outerloop}). Second, to support human decision making, an uncertainty estimation model such as RIO \cite{qiu:iclr20} can be applied to the Predictor, providing confidence intervals around the outcome $O'$. Third, the continual new data collection in the outer loop makes it possible to extend ESP to uncertain environments and to dynamic optimization, where the objective function changes over time \cite{jin2011,Yu2010,jin2005}. By giving higher priority to new examples, the Predictor can be trained to track such changing objectives. Fourth, in some domains, such as those in financial services and healthcare industries that are strongly regulated, it may be necessary to justify the actions explicitly. Rather than evolving a Prescriptor as a neural network, it may be possible to evolve rule-set representations \cite{hodjat2018pretsl} for this role, thus making the decision policy explainable. Such extensions build upon the versatility of the ESP framework, and make it possible to incorporate the demands of real-world applications.

Although the Predictor does not have to be perfect, and its approximate performance can even lead to regularization as was discussed in Section~\ref{sc:regularization}, it is sometimes the bottleneck in building an application of ESP.  In the non-sequential case, the training data may not be sufficiently complete, and in the sequential case, it may be difficult to create episodes that run to successful conclusion early in the training. Note that in this paper, the Predictor was trained with targets from discounted rewards over time. An alternative approach would be to incrementally extend the time horizon of Predictors by training them iteratively \cite{riedmiller2005neural}. Such an approach could help resolve conflicts between Q targets, and thereby help in early training. Another approach would be to make the rewards more incremental, or evolve them using reward function search \cite{houthooft2018evolved,niekum2010genetic}.  It may also be possible to evaluate the quality of the Predictor directly, and adjust sampling from the real world accordingly \cite{jin2003}.

An interesting future application of ESP is in multiobjective domains. In many real-world decision-making domains, there are at least two conflicting objectives: performance and cost. As an evolutionary approach, ESP lends itself well to optimizing multiple objectives \cite{coello:ijkais99,emmerich:natcomp18}. The population forms a Pareto front, and multiple Prescriptors can be evolved to represent the different tradeoffs. Extending the work in this paper, it would be illuminating to compare ESP in such domains to recent efforts in multiobjective RL \cite{liu:ieeetsmc15,yang:neurips19,mossalam:arxiv16}, evaluating whether there are complementary strengths that could be exploited.

\section{Conclusion} 

ESP is a surrogate-assisted evolutionary optimization method designed specifically for discovering decision strategies in real-world applications. Based on historical data, a surrogate is learned and used to evaluate candidate policies with minimal exploration cost. Extended into sequential decision making, ESP is highly sample efficient, has low variance, and low regret, making the policies reliable and safe. Surprisingly, the surrogate also regularizes decision making, making it sometimes possible to discover good policies even when direct evolution fails. ESP is therefore a promising approach to improving decision making in many real world applications where historical data is available.
 
\small
\bibliographystyle{ACM-Reference-Format}
\bibliography{paper}

\end{document}